\documentclass[10pt,twocolumn,letterpaper]{article}

\usepackage[pagenumbers]{cvpr} 

\usepackage{graphicx}
\usepackage{amsmath}
\usepackage{amssymb}
\usepackage{booktabs}

\usepackage[pagebackref,breaklinks,colorlinks]{hyperref}

\usepackage[capitalize]{cleveref}
\crefname{section}{Sec.}{Secs.}
\Crefname{section}{Section}{Sections}
\Crefname{table}{Table}{Tables}
\crefname{table}{Tab.}{Tabs.}

\def\cvprPaperID{7960} 
\def\confName{CVPR}
\def\confYear{2023}

\begin{document}

\title{Flow-guided Semi-supervised Video Object Segmentation}

\author{
Yushan Zhang
\and
Andreas Robinson
\and
Maria Magnusson
\and
Michael Felsberg
}
\maketitle

\begin{abstract}
We propose an optical flow-guided approach for semi-supervised video object segmentation. Optical flow is usually exploited as additional guidance information in unsupervised video object segmentation. However, its relevance in semi-supervised video object segmentation has not been fully explored. In this work, we follow an encoder-decoder approach to address the segmentation task. A model to extract the combined information from optical flow and the image is proposed, which is then used as input to the target model and the decoder network. Unlike previous methods where concatenation is used to integrate information from image data and optical flow, a simple yet effective attention mechanism is exploited in our work. Experiments on DAVIS 2017 and YouTube-VOS 2019 show that by integrating the information extracted from optical flow into the original image branch results in a strong performance gain and our method achieves state-of-the-art performance.
\end{abstract}

\section{Introduction}
\label{sec:intro}

Video object segmentation has been an active research area in the field of computer vision with applications in video editing~\cite{62}, autonomous driving~\cite{63}, and intelligent surveillance~\cite{64}. The field addresses mainly two branches of the problem, semi-supervised video object segmentation~\cite{44} and unsupervised video object segmentation~\cite{10}. The former aims to segment the target object with pixel-wise labels given annotations in the first frame. The latter attempts to segment the object of interest automatically without any ground-truth labels. This work focuses on the semi-supervised video object segmentation problem.

Among the methods to solve semi-supervised video object segmentation, discriminative video object segmentation~\cite{7, 8, 9} works well regarding accuracy and efficiency. Such methods usually follow an encoder-decoder framework. An online few-shot learner is introduced to learn a target model for the objects of interest. The target model is then utilized in the subsequent frames to get the segmentation. It has been shown recently that discriminative methods achieve comparable results with memory-based methods~\cite{2, 3, 4, 5}, but have lower computational requirements, and the possibility to operate in real-time. The previously mentioned methods use RGB frames as input, whereas optical flow~\cite{37, 38, 39, 60} is an essential part in constructing unsupervised video object segmentation algorithms~\cite{10, 11, 12, 13}, it is largely neglected in semi-supervised video object segmentation. We postulate that it is advantageous to integrate optical flow information also into semi-supervised video object segmentation.

\begin{figure}
\begin{center}
\includegraphics[width=0.45\textwidth]{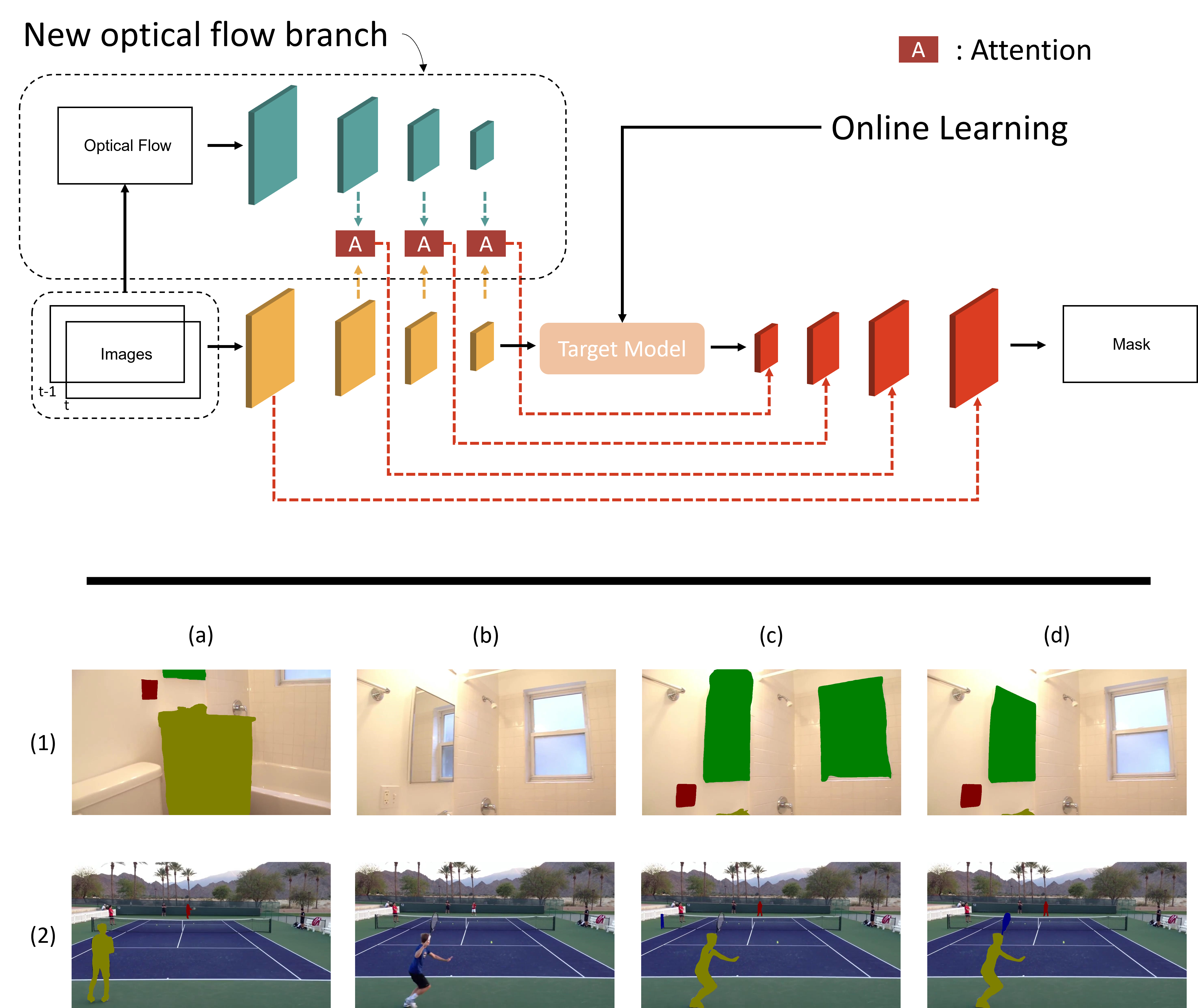}
\end{center}
  \caption{Top: The inference network of the proposed model. We propose to add an optical flow branch to learn a more powerful target model representation and decoder representation. Bottom: Example results on the Youtube-VOS dataset. Rowa (1) and (2) show two different sequences. Columns (a), (b), (c), and (d) are the first frame with annotation, the test frame, the result of the baseline, and the result of the proposed method. In the first example the proposed method managed to distinguish between the mirror and the window with similar appearances. In the second example the proposed method succeed to segment the fast moving tennis racket.}
\end{figure}

In this work, we follow the discriminative video object segmentation paradigm~\cite{7, 8}, and feed the few-shot learner and the decoder network with both image data for appearance features and optical flow for motion features. Figure 1 shows the inference network of the proposed model together with some example results in comparison to our baseline. We choose LWL~\cite{8} as baseline as it is by orders faster than other methods,  such as PReMVOS~\cite{56}. Additionally, to integrate the complementary information from the optical flow, an attention mechanism is utilized to integrate appearance features and motion features. Experiments show that the resulting target model has a higher discriminative power than the baseline. We conduct comprehensive experiments on DAVIS 2017 and YouTube-VOS 2019 datasets to demonstrate the effectiveness of the proposed method.

To summarize, we make the following contributions:

1. We propose to use optical flow as additional information in semi-supervised video object segmentation.

2. A simple yet effective attention mechanism is used to integrate appearance features and motion features for a more discriminative target representation.

3. Extensive experiments on DAVIS 2017 and YouTube-VOS 2019 datasets are conducted to verify that the results of the proposed method are superior to the baseline and achieves state-of-the-art performance.


\section{Related work}
\begin{figure*}[t]
\begin{center}
\includegraphics[width=0.95\textwidth]{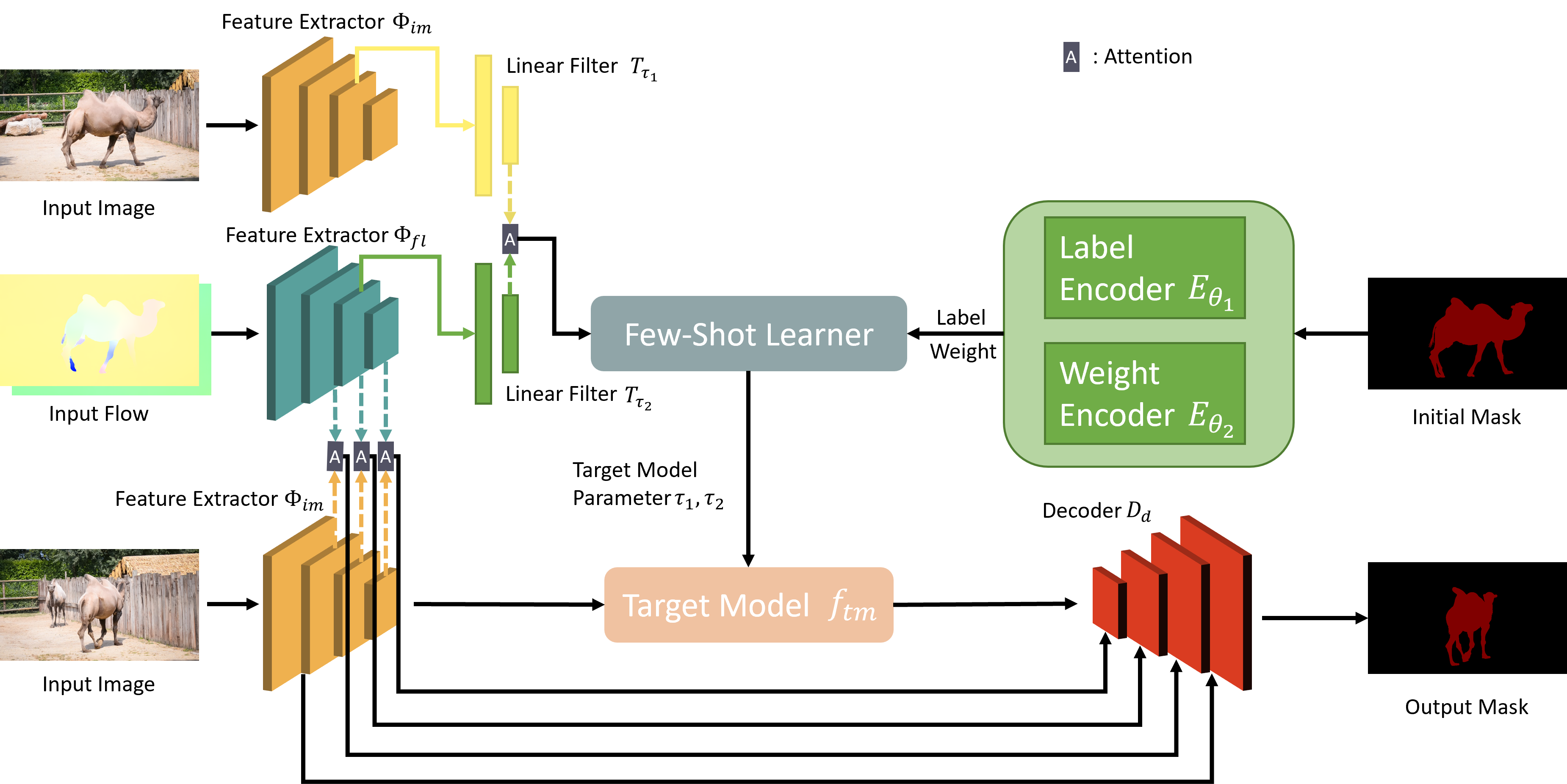}
\end{center}
   \caption{Overall framework of the proposed method. The algorithm consists of two stages. In the first stage, the initial frame, the optical flow, and the ground-truth mask are fed into a few-shot learner to learn the initial parameters of the target model. In the second stage, the target model is applied to subsequent frames and associated optical flow fields to estimate the object representation. The object representation and the features extracted from the current frame and the optical flow are then sent to the decoder network to get the final segmentation mask. Dashed lines indicate the proposed attention network to integrate image features and motion features.}
\end{figure*}
\paragraph{Semi-supervised video object segmentation}
usually takes an encoder-decoder framework, such as memory-based video object segmentation~\cite{2, 3, 4, 5, 6} and discriminative video object segmentation~\cite{7, 8, 9}. 

Space-time memory networks (STM)~\cite{2} encode the image and the predicted mask from all the previous frames and retrieve information from the encoded memory and the current frame. Seong \textit{et al}.~\cite{3} extended STM by introducing kernelized memory network to additionally introduce memory-to-query matching to the framework. To improve the efficiency of STM, Cheng \textit{et al}.~\cite{4}, Li \textit{et al}.~\cite{5}, and Wang \textit{et al}.~\cite{6} modified the memory networks and reduced the computational cost. 

Robinson \textit{et al}.~\cite{7} learnt and updated a discriminative target model from the previous frames, and applied it in the current frame to provide a guidance to the decoder. Bhat \textit{et al}.~\cite{8} extended the work by a label generator to automatically learn from the ground-truth label and provide a high-level guidance to the decoder. Robinson  \textit{et al}.~\cite{9} introduced another prediction branch for similar objects and adjusted the model to be aware of distractors.
\paragraph{Unsupervised video object segmentation}
aims to segment objects from video sequences automatically. In recent years, this is usually solved by taking optical flow as an additional input to learn from motion features ~\cite{10, 11, 12, 13}. 
Fragkiadaki \textit{et al}.~\cite{10} first integrated optical flow and static boundaries together to get segment proposals and then ranked the segment proposals by a Moving Objectness Detector. Tokmakov \textit{et al}.~\cite{11} proposed an encoder-decoder network which takes ground-truth optical flow as input and output motion segmentation. Ren \textit{et al}.~\cite{12} proposed a reciprocal transformation network to correlate three key factors: the intra-frame contrast, the motion cues, and the temporal coherence of recurring objects. Yang \textit{et al}.~\cite{13} introduced a variant of transformer to segment the optical flow into foreground and background, which then serves as a ground-truth for segmentation. 
\paragraph{Optical Flow}
is a vector field between two images, showing how the pixels of an object in the first image can be moved to form the same object in the second image. Deep learning-based optical flow estimation has gained increasing attention in recent years. It can either be trained in a supervised manner~\cite{37, 38} or an unsupervised manner~\cite{39}. Liu \textit{et al}.~\cite{39} distilled reliable optical flow information from non-occlusion areas as ground-truth, and used temporal information from multiple frames to learn optical flow in an unsupervised manner. Sun \textit{et al}.~\cite{37} utilized pyramid processing and warping strategy to construct a cost volumn to estimate the optical flow. Teed \textit{et al}.~\cite{38} introduced a 4D correlation volume and learnt optical flow in a recurrent refinement manner. Researches in video recognition~\cite{40}, object segmentation~\cite{10, 11, 12, 13}, and object tracking~\cite{59, 61} demonstrate that optical flow works well as auxiliary information in vision tasks.
\paragraph{Transformers}
in vision~\cite{14} comes from language tasks~\cite{15}, which work in a self-attention manner to capture the long-term information between input elements. Variations of transformers have been proven to be highly efficient in different research areas of computer vision: image recognition~\cite{18, 19, 20, 21, 22, 23}, object detection~\cite{24, 25}, video object segmentation~\cite{2, 3, 4, 5, 6}, semantic segmentation~\cite{17, 35, 36}, image generation~\cite{26, 27, 28}, low-level vision~\cite{32, 33, 34}, and video understanding~\cite{16, 29, 30, 31}. We propose a variant of transformers on multiple modalities appearance features and motion features. Our proposed method enables channel attention between appearance features and motion features, resulting in a representation with improved discriminative power.


\section{Approach}
In this work, we follow the encoder-decoder paradigm and propose an optical flow-guided video object segmentation method. This is achieved by a two-stage learning framework, where we first learn a target model from the image, optical flow, and the given ground-truth mask in the first frame. The target model is then applied to the subsequent frames to get the object representation. The object representation together with the extracted feature of the current image and optical flow are then fed into the decoder network to generate the final predicted mask. The target model is then updated according to the current frame and the predicted mask. The overall architecture of the proposed method is shown in Figure 2. 
\subsection{Baseline method}
We chose learning what to learn (LWL)~\cite{8} as our baseline method, because it is one the state-of-the-art methods and works well regarding both accuracy and efficiency. The method consists of several key components: the label encoder, the importance weight generator, the target model, and the decoder network.
A label encoder is introduced to automatically learn important information from the ground-truth segmentation mask. As opposed to methods without the label encoder, e.g.~\cite{7}, which only learn a coarse mask from the label, the introduction of the label encoder enables the possibility to learn high-level encoded information from the label. Hence, it gives a richer representation of the target model.
An importance weight generator is introduced due to the imbalance of the training data~\cite{8}. That is, for the ground-truth annotation of each frame, the number of pixels that belong to the foreground and the background are usually not in the same order of magnitude. 
The label encoder and the importance weight generator both take the ground-truth segmentation mask as input, and share the same network architecture. 
The target model is introduced to learn the object representation. For each encoded label channel, two layers of linear filters are learnt online by a few-shot learner.
The decoder network~\cite{41} recovers the mask from the image features of the current frame given the object representation output from the target model.
During training, a training sample of four random frames is generated from one training sequence. One of the frames is used as reference frame for online training of the target model. The other frames are used for offline training of the decoder network.
During inference, the target model is learnt given the first frame and the associated optical flow computed from the first two frames, and then fed into the decoder network in subsequent frames. The output is then used as a ground-truth mask to update the target model.

\subsection{Introduction of optical flow}
We introduce an additional optical flow branch both in the target model and the decoder network. In the target model, for each label channel, the target model learns two filters for image data and optical flow,
\begin{equation}
	f_X=T_{\tau_1}(\phi_{im}^{l_3}(X)),
\end{equation}
\begin{equation}
	f_F=T_{\tau_2}(\phi_{fl}^{l_3}(F)),
\end{equation}
where $X$ and $F$ are the input image data and the optical flow, respectively. $\phi$ is the ResNet-50 backbone network to extract features. $T_{\tau_1}$ and $T_{\tau_2}$ are the linear filters with parameter $\tau_1$ for image data and parameter $\tau_2$ for optical flow. The third layer of the extracted features are utilized as the input to the linear filters $T_{\tau_1}$ and $T_{\tau_2}$. The outputs are then fused to get a multi-modal feature representation,
\begin{equation}
	f_{tm}=\mathcal{F}(f_X, f_F),
\end{equation}
where $\mathcal{F}(\cdot)$ is the feature fusion function. Image feature $f_X$ and flow feature $f_F$ are the outputs of the two-layer linear filters in (1) and (2). $f_{tm}$ is the multi-modal feature representation of the target model.

In the decoder network, motion information is introduced by integrating optical flow backbone features into image backbone features. The features are first extracted by pre-trained ResNet-50 networks $\phi_{im}$ for the images and $\phi_{fl}$ for the optical flows,
\begin{equation}
	b_X^{l_k}=\phi_{im}^{l_k}(X),
\end{equation}
\begin{equation}
	b_F^{l_k}=\phi_{fl}^{l_k}(F),
\end{equation}
then the feature fusion function is applied on different feature layers $l_k$ of the extracted features to integrate the image feature and the motion feature. Specifically, we integrate the features from layer4, layer3, and layer2,
\begin{equation}
  f_{d}^{l_k}=
    \begin{cases}
      \mathcal{F}(b_X^{l_k}, b_F^{l_k}) & \text{k=2,3,4}\\
      b_F^{l_k} & \text{otherwise}\\
    \end{cases} 
\end{equation}
where $l_k$ denotes different layers. The output features from the attention network then serve as the input of the decoder network. We adopt exactly the same decoder network architecture as LWL~\cite{8}.

\begin{figure}
\begin{center}
\includegraphics[width=0.45\textwidth]{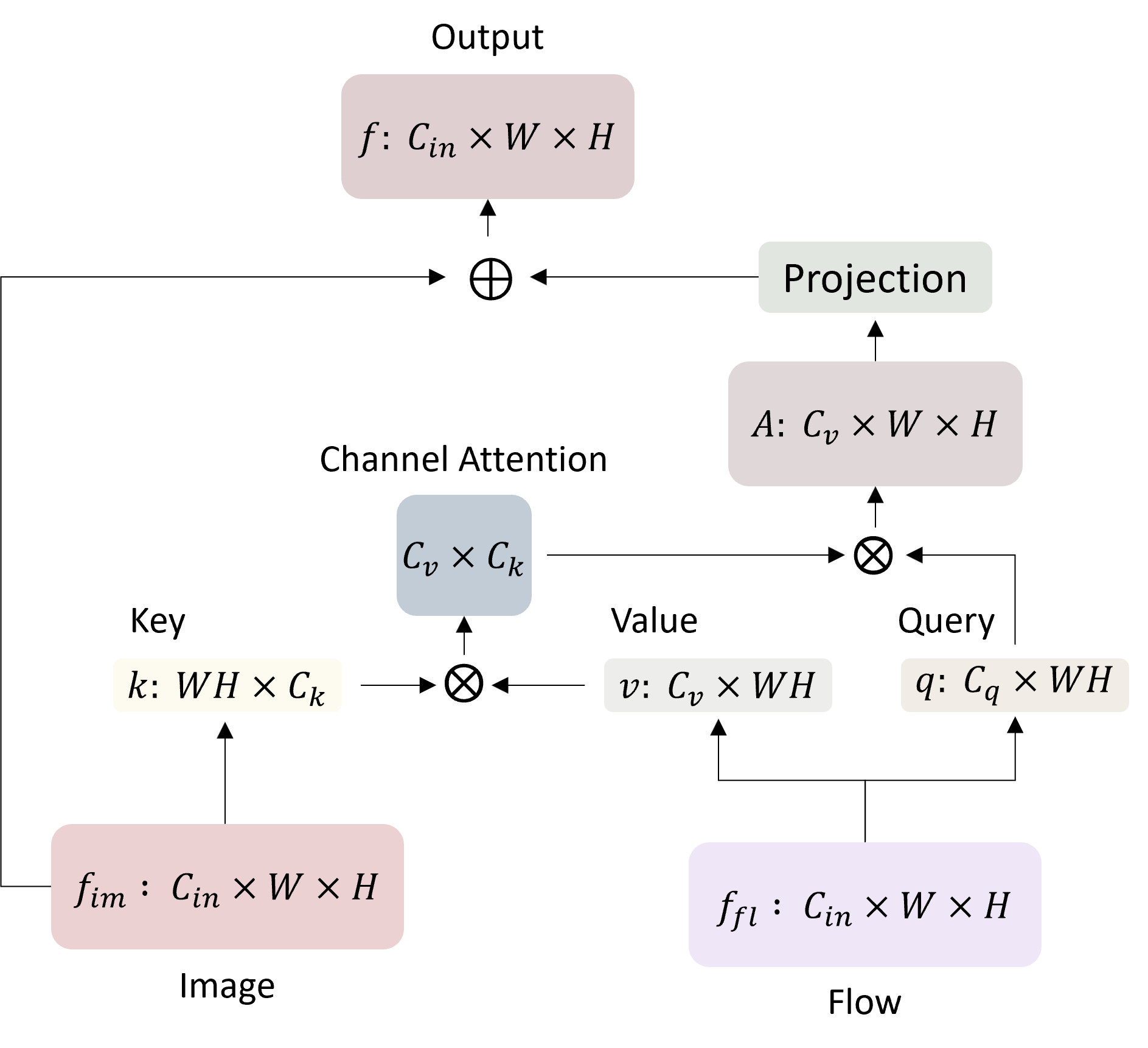}
\end{center}
  \caption{Overall framework of the attention network. The inputs to the network are the image features $f_{im}$ and the optical flow features $f_{fl}$. The outputs of the network $f$ are the multi-modal feature representations obtained. $C_{in}$ is the number of channels of the input features. $W$ and $H$ are the width and height of the input feature. $C_q$, $C_k$ and $C_v$ are the number of channels of the query, key and value respectively. Then Channel attention map has a size of $C_v \times C_k$, which remaps the query in the channel dimension, where $C_q=C_k$. The remapped query is then projected to have the same dimension as the input features to get the output. A skip connected from the input image features is added to the output.}
\end{figure}

\subsection{Multi-modal attention}
To learn the feature representation with both appearance information and motion information, a multi-modal attention is introduced to integrate the appearance feature and the motion feature,
\begin{equation}
	f=A_{\alpha}(f_{im}, f_{fl}),
\end{equation}
where $\alpha$ is the parameter of the attention network. The architecture of the attention network is detailed in Figure 3. Image feature $f_{im}$ and flow feature $f_{fl}$ are the input features to be combined. The query and value are generated from the optical flow feature, the key is generated from the image feature. The feature fusion for the multi-modal feature representation in (3) for the target model and in (6) for the decoder feature can be written as:
\begin{equation}
	f_{tm}=A_{\alpha_{tm}}(f_X, f_F),
\end{equation}
\begin{equation}
  f_{d}^{l_k}=
    \begin{cases}
      A_{\alpha_d^k}(b_X^{l_k}, b_F^{l_k}) & \text{k=2,3,4}\\
      b_F^{l_k} & \text{otherwise}\\
    \end{cases} 
\end{equation}
where $\alpha_{tm}$ is the attention network parameter for the target model. $\alpha_d^k$ is the parameter for the $k$th layer of the decoder backbone features.

\subsection{Optical flow representation}
In deep networks, optical flow vectors are typically embedded into three dimensions with horizontal component, vertical component, and the magnitude of the vectors~\cite{59, 61}. The optical flow elements lie on a cone with the embedded z axis as symmetry axis. However, this representation of optical flow suffers from one problem: values above and below zero represent opposite directions of the flow vector, whereas, for images, values are always non-negative representing the intensity of the input. This difference in nature can be problematic when the optical flow is fed directly into the ResNet backbone. To solve this problem, we pre-process the optical flow representation to all-positive values by first scale the magnitude by a factor of $\sqrt{2}$ and then rotate the cone symmetry axis. We apply the matrix:
\begin{equation}
    R = 
    \begin{bmatrix}
        \frac{1}{2\sqrt{3}}+\frac{1}{2} & \frac{1}{2\sqrt{3}}-\frac{1}{2} & \frac{\sqrt{2}}{\sqrt{3}}\\
        \frac{1}{2\sqrt{3}}-\frac{1}{2} & \frac{1}{2\sqrt{3}}+\frac{1}{2} & \frac{\sqrt{2}}{\sqrt{3}}\\
        -\frac{1}{\sqrt{3}} & -\frac{1}{\sqrt{3}} & \frac{\sqrt{2}}{\sqrt{3}}
    \end{bmatrix}.
\end{equation}
Before the transformation, the sign of the first two channels of the representation encoded flow vector directions in four quadrants. After the rotation, the magnitude of the three channels of the representation encodes the flow vector similar to RGB intensities.

\subsection{Online training}
The parameters for the linear filters in the target model are learnt by a few-shot learning during online training. The image data and the optical flow are first passed through a feature extractor backbone ResNet-50. Separate filters are then applied on the extracted features to get the appearance representation and motion representation. Multi-modal attention (8) is applied on both representations to get the target model. Given the encoded label and the importance weight, the linear filters are then optimized by the following loss function, 
\begin{equation}
\begin{split}
	L_{tm}=\frac{1}{2}\sum_{x_t,y_t\in\mathcal{D}} \parallel E_{\theta_2}(y_t)\cdot(f_{tm}
	-E_{\theta_1}(y_t)) \parallel ^2\\ +\frac{\lambda}{2}(\parallel\tau_1\parallel^2+\parallel\tau_2\parallel^2),
\end{split}
\end{equation}
where $\mathcal{D}$ denotes the training dataset for the few-shot learner. $y_t$ is the ground-truth mask labels in the training dataset. $f_{tm}$ is the target model from (8). For the target model, the third layer of the extracted feature is first utilized as input to the linear filters. Unlike LWL~\cite{8}, we have two linear filters for each label channel, one for image data and one for optical flow. An attention network is then applied on the output of the image linear filter and the optical flow linear filter to get the target representation. The encoder network $E$ with parameter $\theta_1$ gives the encoded label, and with parameter $\theta_2$ gives the importance weight. The learned linear filters, together with attention network are then applied on the following frames to get the target model.

Note that only the two linear filters $T_{\tau_1}$ and $T_{\tau_2}$ are optimized during the online training by Gauss–Newton algorithm. The parameters $\tau_1$ and $\tau_2$ are then updated by the following equation:
\begin{equation}
	\tau^{n+1} = \tau^n - (J_r^TJ_r)^{-1}J_r^Tr(\tau^n),
\end{equation}
 where $r$ is the residual of the loss function in (9), and the Jacobian matrix $J_r$ is obtained from Pytorch autograd tools. 
\subsection{Off-line training}
The decoder network has the same architecture as LWL~\cite{8}, and is trained during off-line training. One training sample consists of four random frames in one video, with one frame as reference frame to train the target model linear filters and three frames as test frames to train the other parts of the network. We first get the target model parameters from the reference frame. Then we use the extracted image features, optical flow features and the target model as input of the decoder network to estimate the final mask. The total training loss function is given by
\begin{equation}
\begin{split}
	L_{tot}=\frac{1}{N-1}\sum_{t=1}^{N-1}\mathcal{L}(D_d(f_{tm},	f_{d}^{l_k}),y_t),
\end{split}
\end{equation}
where $N$ is the length of a training sample. $D$ parameterized by $d$ denotes the decoder network applied on the learnt feature and the target model. $\mathcal{L}(\cdot,\cdot)$ computes the segmentation loss between the prediction and the ground-truth. The two inputs to the decoder are the target representation $f_{tm}$ in (8) learnt by the few-shot learner and the backbone feature $f_{d}^{l_k}$ in (9) that combined the appearance backbone features and the motion backbone features. The final loss is the sum over all training samples.
\subsection{Inference}
During inference, the few-shot learner is optimized on the first frame given the ground-truth label, the first frame, and the optical flow computed from the first two frames. The learned parameters $T_{\tau_1}$ and $T_{\tau_2}$ of linear filters, and the attention network are then applied to the new incoming frames to get the target representation. The target representation and the multi-modal backbone feature are then fed into the decoder network to get the output mask:
\begin{equation}
\begin{split}
	S=D_d(f_{tm}, f_{d}^{l_k})).
\end{split}
\end{equation}
The predicted mask is then used as ground-truth label for the current frame to update the target model.


\section{Experiments}
In this section, we first give the implementation details of our training process. We then conduct an ablation study to verify the effectiveness of the proposed method. The comparison to state-of-the-art is given after that, showing that our method is competitive on YouTube-VOS~\cite{42} and DAVIS~\cite{43} datasets.
\subsection{Implementation details}
Our model is trained on YouTube-VOS~\cite{42} and DAVIS~\cite{43} datasets. Optical flow is obtained from the RAFT~\cite{38} model pre-trained on FlyingChairs and FlyingThings datasets. Following the work of Gladh \textit{et al}.~\cite{59,61}, the optical flow is first pre-processed into three channels as described in Section 3.4. Random flipping, random affine transformation, and color jittering are employed to the first frame for date augment, with the optical flow changed accordingly, before training the few-shot learner. Cropped regions with size $832\times480$ are extracted from the frames and the optical flow. The linear filters in the target model are optimized by steepest gradient descent. Other parts of the model are optimized by Adam optimizer.

We generate our training sample as four random frames and their corresponding optical flows from one sequence. The optical flow is obtained from the sampled frame and its previous frame. If the first frame of the video sequence is sampled, the optical flow is obtained from the first frame and the second frame. We define a frame set as a pair consisting of one image frame and its associated optical flow. One of the frame sets is used as reference frame to train the few-shot learner to get the parameters of the target model. Data augmentation is applied before training the few-shot learner. The other frame sets are then used to train the decoder network. The feature extractor backbone ResNet-50 is initialized with the Mask R-CNN~\cite{52} weights, which is trained particularly for segmentation tasks. The other parts of the network are initialized by He initialization~\cite{53}, which is widely used in convolution neural network.

\subsection{Datasets}
\paragraph{DAVIS} is a densely annotated video segmentation dataset for tasks of Semi-supervised VOS, Interactive VOS, and Unsupervised VOS~\cite{43}. There are two versions of the dataset: DAVIS 2016 with 20 videos for validation, where only one instance is annotated in each video sequence, and DAVIS 2017 with 30 videos for validation, where multiple instances are annotated. There are two metrics to evaluate the video object segmentation performance. To evaluate the region similarity, the $J$ metric is introduced. To evaluate the contour accuracy, the $F$ metric is introduced. For the overall performance, $J\&F$ is defined as the average score of the region similarity $J$ and the contour accuracy $F$.
\paragraph{YouTube-VOS} is a large-scale benchmark for video object segmentation including tasks of Semi-supervised VOS, Video instance segmentation, and Referring VOS~\cite{42}. Overall it has 5000+ high-resolution YouTube videos, 90+ semantic categories, 7800+ unique objects, 190k+ high-quality manual annotations, and 340+ minutes duration. The evaluation metrics are the same as DAVIS dataset, $J$ for region similarity and $F$ for contour accuracy. For the validation set, the objects classes in the video are divided into seen and unseen object categories depending on whether they are seen during the training process. YouTube-VOS evaluation tool compute the two evaluation metrics separately for seen object categories $J_s$, $F_s$ and unseen object categories $J_u$, $F_u$.

\subsection{Ablation study}
In this section, we conduct an ablation study on the introduction of the optical flow guidance to verify the effectiveness of the proposed model on the YouTube-VOS dataset and the DAVIS dataset.

\begin{table}
  \centering
  \begin{tabular}{@{}lccc@{}}
    \toprule
    Method & $J\&F$ & $J$ & $F$ \\
    \midrule
    Baseline & 81.6 & 79.1 & 84.1\\
    Concatenated & 82.2 & 79.0 & 85.5\\
    Ours & \textbf{82.8} & 79.4 & 86.1\\
    \bottomrule
  \end{tabular}
  \caption{Ablation study on DAVIS 2017. $J$ and $F$ represent the mean Jaccard J and boundary F scores. $J\&F$ denotes the overall score of $J$ and $F$. The number of label channels is set as 16.}
  \label{tab:table1}
\end{table}

\begin{table}
  \centering
  \begin{tabular}{@{}lccccc@{}}
    \toprule
    Method & $J\&F$ & $J_{s}$ & $J_{u}$ & $F_{s}$ & $F_{u}$ \\
    \midrule
    Baseline & 81.0 & 79.6 & 76.4 & 83.8 & 84.2\\
    Concatenated & 80.9 & 79.5 & 76.1 & 83.8 & 84.2\\
    Ours & \textbf{81.7} & 79.6 & 77.7 & 83.8 & 85.7\\
    \bottomrule
  \end{tabular}
  \caption{Ablation study on YouTube-VOS 2019.  $J_s$ and $J_u$ represent the mean Jaccard of the object categories that are seen and unseen during training. $F_s$ and $F_u$ represent the boundary scores of the object categories that are seen and unseen during training. $J\&F$ denotes the overall score of $J$ and $F$. The number of label channels is set as 16.}
  \label{tab:table2}
\end{table}

We compare three models: the baseline model without the optical flow input branch (Baseline), the proposed model with the optical flow input branch (Ours), and the model that takes the concatenated image and optical flow feature as input (Concatenated). For the label encoder we set the number of channels as 16. Overall results on both datasets are given in Table 1 and Table 2. The results of the baseline method and the concatenated method on both dataset are almost the same. By applying a simple channel attention to combine the appearance feature and the motion feature, the performance is enhanced. The proposed method improves the overall performance of DAVIS 2017 by 1.2 percent points and YouTube-VOS 2019 by 0.7 percent points. For the DAVIS 2017 dataset, the region similarity $J$ and contour accuracy $F$ improves by 0.3 percent points and 2.0 percent points. For the YouTube-VOS 2019 dataset, the region similarity $J$ and contour accuracy $F$ improves by 0.7 percent points and 0.8 percent points. The performance gain on the YouTube-VOS 2019 dataset is almost the same for both metrics. On the DAVIS 2017 dataset, the performance gain of contour accuracy is much higher than region similarity. A possible explanation is that the introduction of the optical flow branch benefits the contour prediction, since it is usually the case that moving objects have a different optical flow field and hence the object boundary is predicted. Also the DAVIS dataset is a more carefully annotated dataset with fine contours compared to the YouTube-VOS dataset~\cite{9}. Moreover, for the YouTube-VOS dataset, the region similarity $J$ and contour accuracy $F$ for seen object categories remain the same as the baseline, and for unseen object categories the improvement is 1.3 percent points and 1.5 percent points, which indicates that the proposed model is less overfitted to the training dataset and has a better generalization ability. A possible explanation is that without motion features, the object segmentation mainly relies on the appearance feature, and the model tends to remember the object classes that have been seen during training. By additionally introducing the motion feature, the model relies less on the appearance features and the accuracy discrepancy between seen and unseen object categories decreases.

\begin{table}
  \centering
  \begin{tabular}{@{}lccc@{}}
    \toprule
    Method & $J\&F$ & $J$ & $F$\\
    \midrule
    OnAVOS~\cite{58} & 67.9 & 64.5 & 70.5\\
    AGAME~\cite{57} & 70.0 & 67.2 & 72.7\\
    FEELVOS~\cite{55} & 71.5 & 69.1 & 74.0\\
    FRTM~\cite{7} & 76.7 & - & -\\
    PReMVOS~\cite{56} & 77.8 & 73.9 & 81.7\\
    LWL~\cite{8} & 81.6 & 79.1 & 84.1\\
    CFBI~\cite{47} & 81.9 & 79.1 & 84.6\\
    SST~\cite{48} & \textcolor{blue}{82.5} & \textcolor{green}{79.9} & \textcolor{blue}{85.1}\\
    Joint~\cite{46} & \textcolor{red}{83.5} & \textcolor{red}{80.8} & \textcolor{red}{86.2}\\
    Ours & \textcolor{green}{82.8} & \textcolor{blue}{79.4} & \textcolor{green}{86.1}\\
    \bottomrule
  \end{tabular}
  \caption{State-of-the-art comparison on DAVIS 2017. Color red, green, blue represent the algorithm that ranked the first, second, and third respectively.}
  \label{tab:table3}
\end{table}

\begin{table}
  \centering
  \begin{tabular}{@{}lcccccc@{}}
    \toprule
    Method & $J\&F$ & $J_{s}$ & $J_{u}$ & $F_{s}$ & $F_{u}$\\
    \midrule
    OnAVOS~\cite{58} & 55.2 & 60.1 & 46.1 & 62.7 & 51.4\\
    PReMVOS~\cite{56} & 66.9 & 71.4 & 56.5 & 75.9 & 63.7\\
    STM-~\cite{2} & 68.2 & - & - & - & -\\
    FRTM~\cite{7} & 72.1 & 72.3 & 65.9 & 76.2 & 74.1\\
    CFBI~\cite{47} & 81.0 & \textcolor{blue}{80.6} & 75.2 & \textcolor{red}{85.1} & 83.0\\
    LWL~\cite{8} & 81.0 & 79.6 & 76.4 & \textcolor{blue}{83.8} & \textcolor{blue}{84.2}\\
    SST~\cite{48} & \textcolor{green}{81.8} & \textcolor{red}{80.9} & \textcolor{blue}{76.6} & - & -\\
    Joint~\cite{46} & \textcolor{red}{82.8} & \textcolor{green}{80.8} & \textcolor{red}{79.0} & \textcolor{green}{84.8} & \textcolor{red}{86.6}\\
    Ours & \textcolor{blue}{81.7} & 79.6 & \textcolor{green}{77.7} & \textcolor{blue}{83.8} & \textcolor{green}{85.7}\\
    \bottomrule
  \end{tabular}
  \caption{State-of-the-art comparison on YouTube-VOS 2019. Color red, green, blue represent the algorithm that ranked the first, second, and third respectively.}
  \label{tab:table4}
\end{table}

\subsection{Comparison to state-of-the-art}
In this section, we give quantitative and qualitative state-of-the-art comparison on the DAVIS 2017 dataset and the Youtube-VOS 2019 dataset.
\subsubsection{Quantitative Comparison}
Table 3 and Table 4 list the performance of our model and state-of-the-art VOS models on the DAVIS 2017 and the YouTube-VOS 2019 datasets. 

We compare our model against state-of-the-art video object segmentation methods. Among which our method showed competitive results on both the YouTube-VOS 2019 dataset and the DAVIS 2017 dataset, the proposed method performs third and second best among all the methods. The results show that the discrepancy of the $J$ score between the seen classes and unseen classes during training is far less than all the other methods. For $F$ score, our proposed method performs even better on unseen classes than on seen classes. The overall performance $J\&F$ of our method are quite even for seen classes (81.7) and unseen classes (81.7), which is never achieved by previous methods. This coincides with our previous hypothesis that the model is less overfitted to the appearance features of the seen classes during training. The results on both datasets demonstrate the effectiveness of the proposed method. 

\begin{figure*}
\begin{center}
\includegraphics[width=0.98\textwidth]{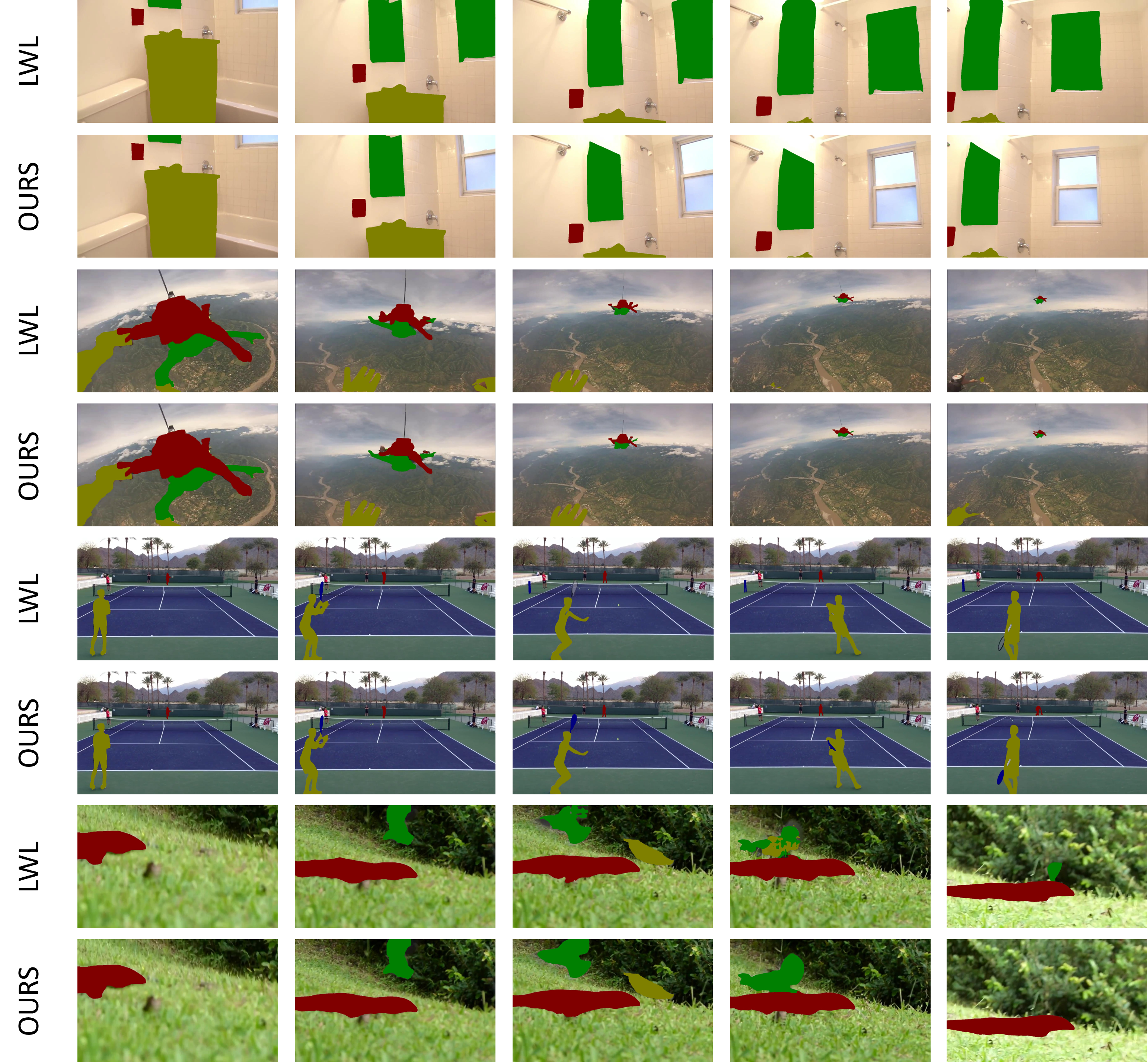}
\end{center}
  \caption{Qualitative results of the proposed model (OURS) and the baseline model (LWL). Four video sequences are selected to demonstrate the effectiveness of the proposed method. The first video include a mirror to be segmented and a window, whose reflection is in the mirror. The second video include object (hand in the bottom left corner) moving out of and into the video scene. In the third video, two persons are playing tennis, with the tennis racket moving very fast. In the fourth video, two magpies are chasing each other and even overlap when they come close to each other.}
\end{figure*}

\subsubsection{Qualitative Comparison}
Figure 4 visualizes qualitative results of the proposed model (OURS) and the baseline model (LWL) on YouTube-VOS dataset. We selected several representative videos with difficulties in the video object segmentation task, such as similar objects, re-identification, and fast motion.

In the first video, the green mask tracks a mirror, in which the window is reflected. LWL failed to distinguish the mirror with the window, hence labeled them both. OURS succeeded to recognize the mirror despite of the similarity between the reflection and the window. This shows that our method is good at distinguishing similar objects. In the second video, the hand in the lower left corner disappears for some frames. LWL lost the target after it disappears. OURS managed to resume tracking after the target returned. This shows that our method is good at re-identification. In the third video, the tennis racket is small in scale and moves really fast. LWL lost the target from the beginning, while OURS kept track of the tennis racket even under severe deformation. This shows that our method is robust under fast motion of small objects. In the fourth video, one magpie is chasing the other. LWL got confused when the two magpies overlap each other and labeled it into two different identities. OURS managed to decide which one is in the front and correctly labeled it as a single object. This shows that our method could also handle overlapping similar objects.


\section{Conclusions}
In this paper, we propose to use optical flow as additional information for semi-supervised video object segmentation. To achieve this purpose, a multi-modal attention mechanism is adopted to integrate appearance features and motion features in the target model and the decoder network to learn a better target representation. We demonstrate the effectiveness of the proposed method by an ablation study on baseline comparison and the number of label channels. By introducing the motion feature, the model relies less on the appearance features and is more robust under circumstances such as similar objects distraction and fast motion. Experiments on DAVIS 2017 and YouTube-VOS 2019 datasets show that the proposed method achieves state-of-the-art performance.


{\small
\bibliographystyle{ieee_fullname}
\bibliography{main}
}

\end{document}